\documentclass{bmvc2k}

\usepackage{siunitx}
\usepackage{amsfonts}
\usepackage{bm}
\usepackage{mathtools}

\title{Order Matters: Shuffling Sequence Generation for Video Prediction}

\addauthor{Junyan Wang}{j.wang81@newcastle.ac.uk}{1}
\addauthor{Bingzhang Hu}{bingzhang.hu@newcastle.ac.uk}{1}
\addauthor{Yang Long}{yang.long@newcastle.ac.uk}{1}
\addauthor{Yu Guan}{yu.guan@newcastle.ac.uk}{1}

\addinstitution{
 Open Lab, School of Computing\\
 Newcastle University\\
 Newcastle Upon Tyne, UK
}
\runninghead{J. Wang \etal}{Shuffling Sequence Generation for Video Prediction}

\def\etal{\emph{et al}\bmvaOneDot}

\newcommand{\fs}{\text{.}}
\newcommand{\com}{\text{,}}
\newcommand{\figref}[1]{Fig.\ref{#1}}
\newcommand{\eqnref}[1]{Eq.\ref{#1}}
\newcommand{\subfigurewidth}{0.5\textwidth}
\DeclarePairedDelimiter\floor{\lfloor}{\rfloor}

\begin{document}

\maketitle

\begin{abstract}
	Predicting future frames in natural video sequences is a new challenge that is receiving increasing attention in the computer vision community. However, existing models suffer from severe loss of temporal information when the predicted sequence is long. Compared to previous methods focusing on generating more realistic contents, this paper extensively studies the importance of sequential order information for video generation. A novel \textit{Shuffling sEquence gEneration network} (SEE-Net) is proposed that can learn to discriminate unnatural sequential orders by shuffling the video frames and comparing them to the real video sequence. Systematic experiments on three datasets with both synthetic and real-world videos manifest the effectiveness of shuffling sequence generation for video prediction in our proposed model and demonstrate state-of-the-art performance by both qualitative and quantitative evaluations. The source code is available at \url{https://github.com/andrewjywang/SEENet} 
\end{abstract}

%-------------------------------------------------------------------------
\section{Introduction}
Unsupervised representation learning is one of the most important problem in the computer vision community. Compared to image, video contains more complex spatio-temporal relationship of visual contents and has much wider applications\cite{zhu2018towards,guan2013robust,long2017zero}. In order to explicitly investigate the learnt representation, video prediction has become an emerging field that can reflect whether temporal information is extracted effectively. There are recent variations of related work on human activity prediction and recognition \cite{lan2014actionpredict,ma2016learningsltm,kong2017deepsequential,guan2014reducing,guan2017ensembles}, motion trajectory forecasting  \cite{alahi2016sociallstm,lee2017desire}, future frame prediction  \cite{mathieu2015gdl,lotter2016prednet} and so on. Also, the application has appeared in robotics  \cite{finn2016robotic} and healthcare \cite{paxton2018healthcare} areas. 

In order to predict long-term future frames, the key challenge is to extract the sequential order information from still contexts. Most of state-of-the-art methods  \cite{lotter2016prednet,mathieu2015gdl} exploited advanced generative neural network to directly predict frames based on reconstruction loss that is more sensitive to contextual information. Some recent works \cite{villegas2017mcnet,denton2017drnet} attempted to extract spatio-temporal information using motion-sensitive descriptors, \textit{e.g.} optical flow, and achieved improved results. A common issue in existing works is the severe loss of temporal information, for example, the target becomes blurry and gradually disappears during processive video prediction.

\begin{figure}
	\centering
	\includegraphics[height=12ex]{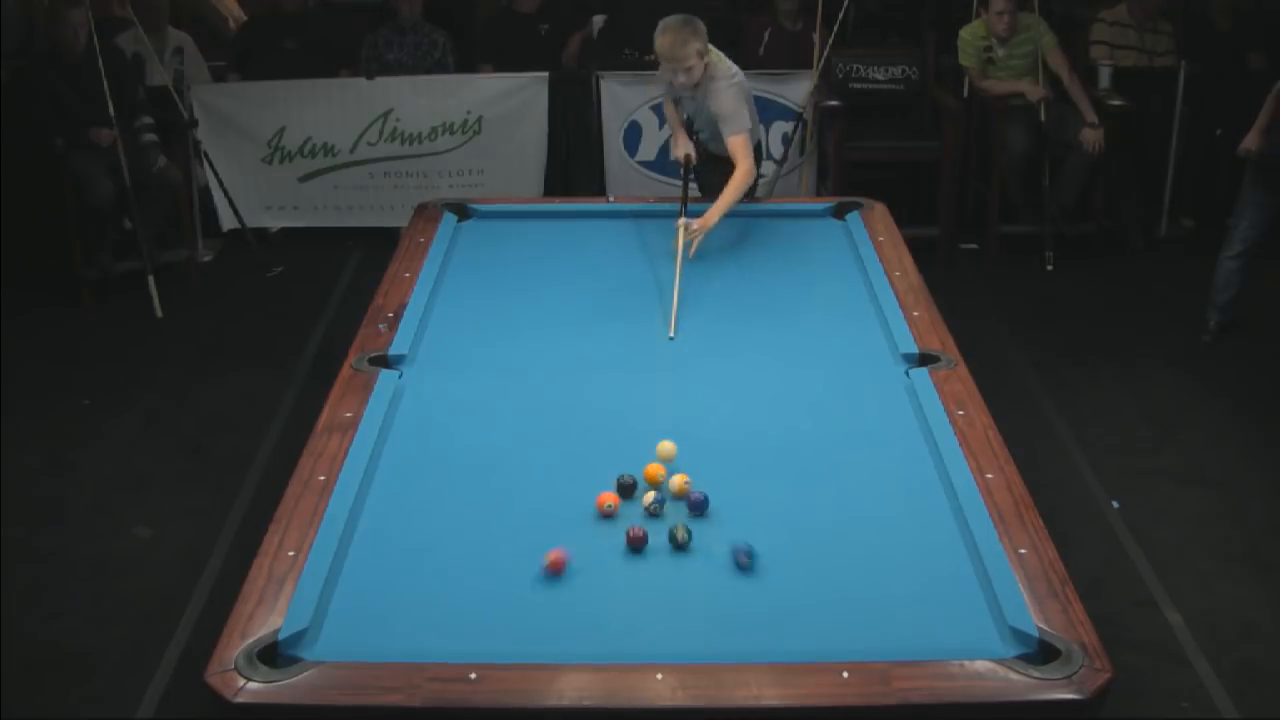}
	\includegraphics[height=12ex]{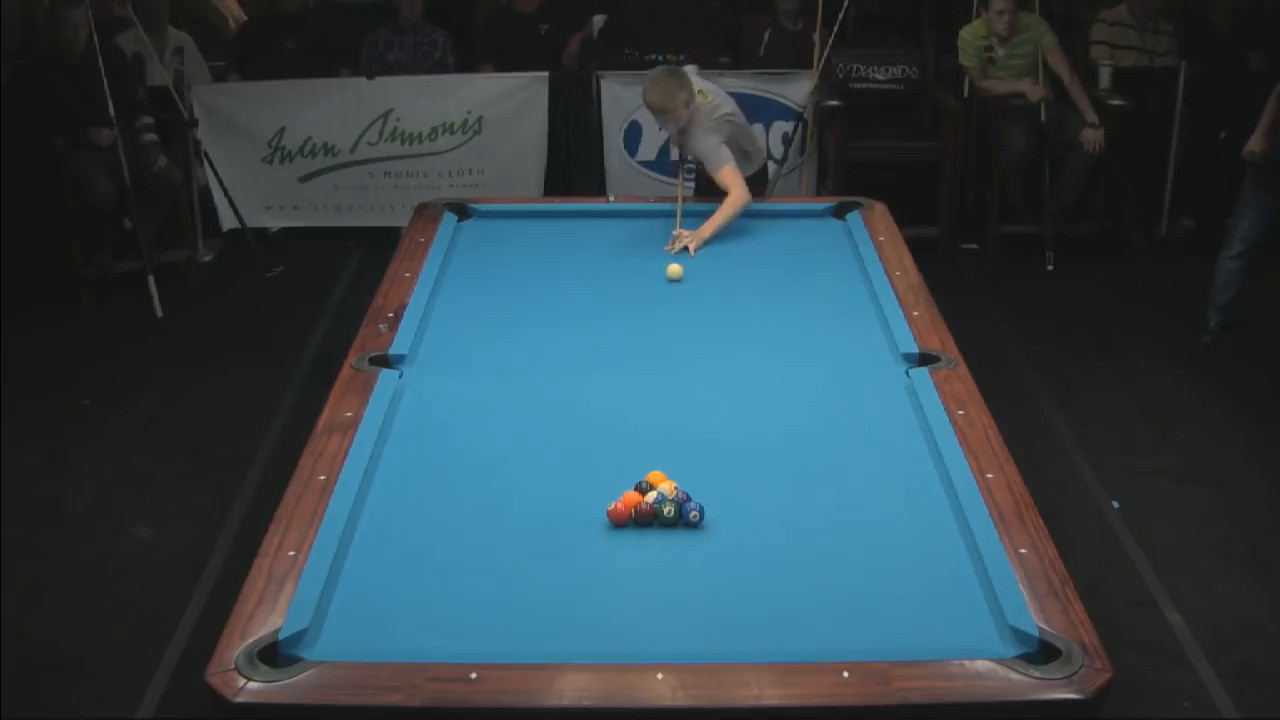}
	\includegraphics[height=12ex]{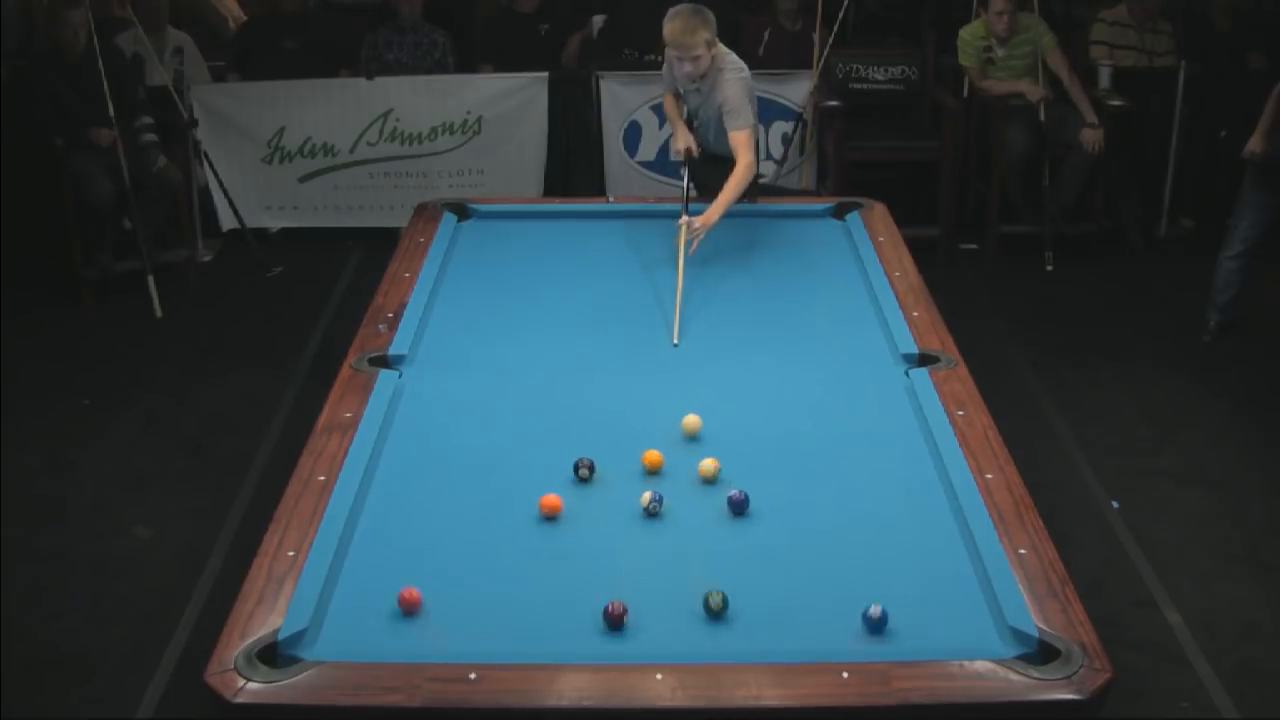}
	\caption{Human can figure out the correct order of shuffled video frames (2-1-3). By doing so, attention is enforced to be paid on the temporal information.\label{fig_1}}
	\vspace{-2ex}
\end{figure}

Our work focuses on future frame prediction and this paper is inspired by a fact that the ordinal information among the frames is more important for the humans' perceptions of a video. And such kind of information can be better captured by performing a sorting task. For example, as shown in \figref{fig_1}, to sort the frames, one has to pay his attention to the temporal information. Motivated by the fact above, we propose a \textit{Shuffling sEquence gEneration network} (SEE-Net) that can explicitly enforce the temporal information extraction by shuffling the sequential order of training videos, where a \textit{Shuffle Discriminator} ($SD$) is designed to distinguish the video sequential with natural  and shuffled order. As the content information is supposed to be the same between real and shuffled frames, the model is therefore forced to extract the temporal order information explicitly. Extracting temporal information is a very challenging task from raw video frames and optical flow is widely used in temporal information extraction tasks \cite{simonyan2014twostream,walker2015denseof,xue2016visualdynamic}, therefore we apply the optical flow network PWCNet  \cite{sun2018pwcnet} to generate optical flow images between adjacent frames. In addition, we evaluate our method on both synthetic dataset (Moving MNIST) and real-world datasets (KTH Actions and MSR Actions). The contributions of this paper are summarized below:
	\begin{itemize}
		\item We propose the SEENet for the task of long-term future frame prediction, which can extract both content and motion information by two independent auto-encoder pathways.
		\item We introduce the shuffle discriminator to explicitly control the extraction of sequential information.
		\item Extensive results manifest that our model is not only more stable in long-term video frame predictions, but also infers more accurate contents at each frame compared to other methods.
	\end{itemize}

\section{Related Work}
\noindent\textbf{Content based Approaches} The task of video frame prediction has received growing attention in the computer vision community. Early work investigates object motion prediction \cite{walker2014usvp}. Advanced neural network approaches were then applied to directly predict future frames \cite{mathieu2015gdl,lotter2016prednet,vondrick2016,vondrick2017}. Mathieu \etal \cite{mathieu2015gdl} proposed a multi-scale auto-encoder network with both gradient difference loss and adversarial loss. PredNet \cite{lotter2016prednet} is inspired by the concept of predictive coding from the neuroscience literature. Each layer in the PredNet model produced local predictions and only forward deviations from these predictions to the subsequent network layers. Vondrick \etal \cite{vondrick2016,vondrick2017} conducts a deep regression network to predict future frame representations. Unlike future frame prediction, Babaeizadeh \etal \cite{babaeizadeh2017sv2p} and Lee \etal \cite{lee2018svap} address the video prediction by stochastic approaches that can predict a different possible future for each sample of its latent variables. A shared drawback of these methods is lack of explicit control of temporal information extraction, and therefore our work disentangles the motion information from video frames to better learn temporal information.

\noindent\textbf{Content-motion Disentanglement} Many recent works use content-motion disentanglement in many ways \cite{xingjian2015convlstm,srivastava2015uselstm,villegas2017mcnet,gao2016representationlearning,zhang2018triple}. For example, Shi \etal \cite{xingjian2015convlstm} proposed a convolutional network which offered a method to obtain time-series information between images and Srivastava \etal \cite{srivastava2015uselstm} demonstrated that Long Short-Term Memory was able to capture pixel-level dynamics by conducting a sequence-to-sequence model. However, direct prediction usually produces blurred images because the reconstruction-based loss function encourages an averaged output. Instead of directly generating pixels, disentangling video into motion and content representation has been widely adopted in recent work. Early disentangled representation \cite{gao2016representationlearning} is originally applied to scene understanding \cite{eslami2016sceneunderstanding} and physics modelling \cite{chang2016physicsmodeling}. MCNet \cite{villegas2017mcnet} disentangles motion from content using image differences and applies the last frame as a content exemplar for video prediction. DrNet \cite{denton2017drnet} disentangles the representation into content and pose information that is penalised by a discrimination loss with encoding semantic information. Other than disentangled approaches, optical flow is the most common approach to extract motion of objects explicitly. For instance, Simonyan \etal \cite{simonyan2014twostream} proposed a two-stream convolutional network for action recognition in videos. Recent works \cite{dosovitskiy2015flownet,ilg2017flownet2,sun2018pwcnet} focus on using convolutional neural network to generate optical flow images and then extended to future frames generation \cite{walker2015denseof,xue2016visualdynamic,patraucean2015spatio}. However, content-motion disentanglement is not sufficient to learn distinct motion information. Therefore, our work also learns ordinal information among the frames.

\noindent\textbf{Shuffle based Self-supervised Learning} Several works utilise shuffle based self-supervised learning methods on videos, which do not require external annotations \cite{wang2015self1,misra2016self2,lee2017self3}. In \cite{wang2015self1}, based on ordinal supervision provided by visual tracking, Wang and Gupta designed a Siamese-triplet network with a ranking loss function to learn the visual representations. Misra \etal \cite{misra2016self2} proposed a self-supervised approach using the convolutional neural network (CNN) for a sequential verification task, where the correct and incorrect order frames are formed into positive and negative samples respectively to train their model. Lee \etal \cite{lee2017self3} presented a self-supervised representation learning approach using temporal shuffled video frames without semantic labels and trained a convolutional neural network to sort the shuffled sequences and output the corrects. In this work, we apply the shuffle based self-supervised learning method on optical flow images to extract the ordinal information from the motion of objects, surfaces, and edges. 

\section{Problem Statement}
To formulate the future frame prediction problem we declare some notations in advance. We use bold lowercase letters $\boldsymbol{x}$ and $\boldsymbol{m}$ to denote one video frame and one optical flow frame, respectively. In particular, the bold lowercase letter $\boldsymbol{h}$ is denoted as the extracted latent feature. We employ the capital letter $E$ to represent the encoding network and the capital letter $G$ to represent the generating network. Besides, we use the curlicue letter $\mathcal{L}$ to represent the loss function.

Formally, given a sequence of $t$ frames from video $\mathbb{X} = \{\boldsymbol{x}_{1}, \boldsymbol{x}_{2}, \dots, \boldsymbol{x}_{t}\}$, we aim to build a model for predicting the next $k$ frames $\{\hat{\boldsymbol{x}}_{t+1}, \hat{\boldsymbol{x}}_{t+2}, \dots, \hat{\boldsymbol{x}}_{t+k}\}$ by learning the video frame representations. The network disentangles the information from input $\boldsymbol{x}$ into time-varying (motion) information $\boldsymbol{h}^m$ and time-invariant (content) information $\boldsymbol{h}^c$ by the motion encoder $E_m$ and content encoder $E_c$ respectively due to the spatial-temporal disentanglement strategy. However, many existing methods only consider the temporal information at the representation level and they \cite{denton2017drnet,hsieh2018ddpae} can not achieve reasonable results on certain scenarios. Inspired by the shuffle based self-supervised learning, the proposed SEE-Net shuffles sequence generation orders and the shuffle discriminator $SD$ converges only if effective temporal information is extracted.
 
\section{SEE-Net}
\begin{figure*}
\centering
  \includegraphics[width=0.9\textwidth]{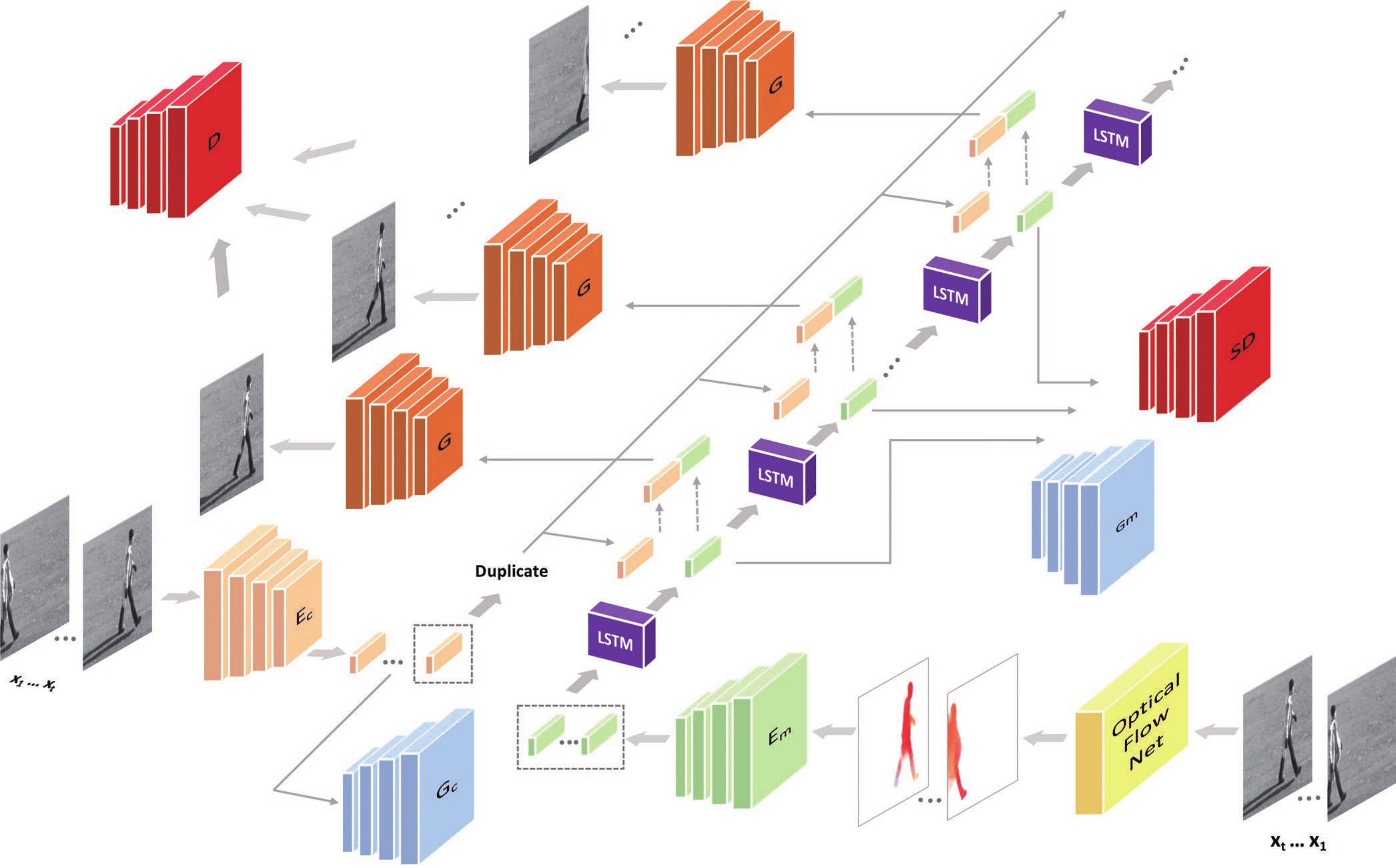}
  \caption{The proposed video prediction framework}
  \label{fig:framework}
\end{figure*}

In this section we introduce the proposed \textbf{SEE-Net} and \figref{fig:framework} illustrates the overall pipeline of the proposed SEE-Net. There are mainly three components in SEE-Net, which includes a motion encoder $E_m$ with decoder $G_m$, a content encoder $E_c$ with decoder $G_c$ as well as a frame generator $G$. In the following, we detail each of them in turn. To explicitly extract the motion information, instead of feeding the original frames directly into $E_m$, we input the optical flow images $\mathbb{M} = \{\boldsymbol{m}_{1}, \boldsymbol{m}_{2}, ...,\boldsymbol{m}_{t-1}\}$ which are extracted between adjacent frames by the advanced optical flow network PWCNet \cite{sun2018pwcnet}. Therefore the motion information can be obtained as $\boldsymbol{h}^m = E_m(\boldsymbol{m})$ and the content information can be obtained as $\boldsymbol{h}^c = E_c(\boldsymbol{x})$. The $E_m$ and $E_c$ are trained in a fashion auto-encoder together with their corresponding decoder $G_m$ and $G_c$.

Subsequently, with the motion information of the previous $t$ frames, we employ a long short-term memory (LSTM) network $f^{lstm}$ to predict the motion information of the next $k$ frames. For short term video prediction, we assume there is no significant background change in the next k frames, and use the content feature $\boldsymbol{h}^c_{t}$ together with the predicted motion feature designed as follows.
\begin{equation}\label{eq:x^T+i}
\begin{split}
	&\boldsymbol{h}^m_{t+i-1} = f^{lstm}(\boldsymbol{h}^m_i, \boldsymbol{h}^m_{i+1}, \cdots , \boldsymbol{h}^m_{i+t-2})\com \\
	&\hat{\boldsymbol{x}}_{t+i} = G([\boldsymbol{h}^c_{t},\boldsymbol{h}^m_{t+i-1}])\com
\end{split}
\end{equation}

where $i=1, \dots, k$ and $[\cdot,\cdot]$ denotes the vector concatenate operation.

\subsection{Content Consistency}
To extract time-invariant content information, the contrastive loss is applied to optimise the content encoder. We use subscripts to denote the indices of different video clips, therefore the odd and even frames in $i^{th}$ video can be denoted as $\boldsymbol{x}^{i}_{2n-1}$ and $\boldsymbol{x}^{i}_{2n}$, where $n = 1, \dots, \floor{\frac{t}{2}}$. It is intuitive that the content information, such as the background and the objects within the same video clip are supposed to be consistent while the content information in different video clips may be various. Thus we have
\begin{equation}\label{eq:L_consistency}
\begin{split}
	\mathcal{L}_{consistency} &= \sum_{i,j} y(\mathcal{D}_{ij})^{2} + (1-y)max(\delta - \mathcal{D}_{ij},0)^{2}\com \\ 
	\text{where}~~\mathcal{D}_{ij} &= \frac{1}{n}\sum_{n} ||E_{c}(\boldsymbol{x}^{i}_{2n-1})-E_{c}(\boldsymbol{x}^{i}_{2n})||_{2}\fs
\end{split}
\end{equation}
$\mathcal{D}_{ij}$ is used to denote the frame-wise $l_2$ distances between content information in $i^{th}$ and $j^{th}$ video clips. We set $y$ as 1 if $i=j$ and 0 otherwise. Therefore minimising $\mathcal{L}_{consistency}$ can ensure $E_c$ to extract similar content information for all frames in the same video clip while at least a $\delta$ difference for those from different clips.

\subsection{Shuffle Discriminator}
To explicitly extract motion information, we propose a novel shuffle discriminator (SD), which takes a sequence of predicted motion information from $f^{lstm}$ as input and discriminates if they are in the correct order. For example, we manually construct two sequences $S_{predicted}$, which is organised in right order; and $S_{shuffled}$, the order of motion information in which is shuffled. The $SD$ consists of a bidirectional LSTM as the first layer and followed by a fully-connected layer.

The $SD$ is supposed to predict 1 for $S_{predicted}$ for its correct order whilst 0 for $S_{shuffled}$. Therefore a \textit{shuffle loss} can be defined as:
\begin{equation}\label{eq:l_shuffle}
	\mathcal{L}_{shuffle} = -log(SD(S_{predicted})) - log(1-SD(S_{shuffled})).
\end{equation}
The shuffle loss forces the predicted motion information $\boldsymbol{h}^m_{t}, \boldsymbol{h}^m_{t+1}, \dots,\boldsymbol{h}^m_{t+k-1}$ to be distinct with each other. Otherwise the SD cannot distinguish the correct ordered sequences from shuffled ones. The $f^{lstm}$ is constrained to learn temporal change between frames thus generate reasonable motion information for upcoming frames. As the correct and shuffled sequences can be constructed without extra labelling, it can be regarded as self-supervised learning.

\subsection{Adversarial Objective for Generator}
Generative adversarial networks (GANs) have yielded promising results in many areas, especially in image generation. Such success is mainly benefited from the adversarial training idea in GANs where the generator is trained competing with the discriminator. In our work, to generate realistic frames and train the generator $G$, we employ the adversarial loss as:
\begin{equation}\label{eq:l_ad}
	\min_{G} \max_{D} \mathcal{L}_{adversarial} = -log(D(\boldsymbol{x}_{t+i}))-log(1-D(G([\boldsymbol{h}^c_{t},\boldsymbol{h}^m_{t+i-1}])))
\end{equation}
To further enhance the quality of the generated frame, inspired by \cite{hu2018}, we employ the $l_1$ loss to minimise the difference between generated frame and ground truth.

\subsection{Optimisation and Training}
Combining the Eq.~\ref{eq:L_consistency}\ref{eq:l_shuffle}\ref{eq:l_ad}, the overall objective can be written as:
\begin{equation}
	\mathcal{L} = \mathcal{L}_{content} + \mathcal{L}_{motion}+\mathcal{L}_{generate} \com
\end{equation}
where
\begin{equation}
\begin{split}
	\mathcal{L}_{content} &= \lambda_1 \mathcal{L}_{consistency} + \lambda_2||\boldsymbol{x}_{j}- G_c(E_c(\boldsymbol{x}_{j}))|| \com \; \textrm{for } t\geq j \geq1\\
	\mathcal{L}_{motion} &= \lambda_3 \mathcal{L}_{shuffle}+\lambda_4 ||\boldsymbol{m}_{t+i-1}- G_m(E_m(\boldsymbol{m}_{t+i-1}))||\com \\
	\mathcal{L}_{generate} &= \alpha \mathcal{L}_{adversarial}+\beta ||\boldsymbol{x}_{t+i}-G([\boldsymbol{h}^c_{t}, \boldsymbol{h}^m_{t+i-1}])||_1 \fs
\end{split}
\end{equation}
The $\lambda_{1-4}$, $\alpha$ and $\beta$ are hyperparameters that control the occupation of each loss. As a framework with multiple loss is difficult to train, in practice, we first optimise the $\mathcal{L}_{content}$ and $\mathcal{L}_{motion}$ respectively until model converges. Then we fixed the weights in $E_m$ and $E_c$ to train the generator $G$ and discriminator $D$. After $G$ and $D$ reaching their optimal, we combine each component and fine-tune the whole network.

\section{Experiments}
Considering video prediction is an early stage problem with various settings, for a fair comparison, we mainly compare with two representative state-of-the-art methods MCNet \cite{villegas2017mcnet} and DrNet \cite{denton2017drnet}. This paper provides a thorough evaluation of the proposed SEE-Net on both a synthetic dataset (Moving MNIST \cite{srivastava2015mnist}) and two simple real-world datasets (KTH Actions, MSR Actions) for many existing work reports they \cite{denton2017drnet, hsieh2018ddpae} cannot achieve reasonable results on certain scenarios.  Both qualitative and quantitative evaluation metrics are adopted to better understand the advantages of our model. 

\noindent\textbf{Model Configuration} Content encoder $E_{c}$, motion encoder $E_{m}$, content decoder $G_{c}$, motion decoder $G_{m}$, and the generator $G$ consist of 4 convolutional layers and two fully connected layers. Each convolutional layer is followed by instance normalization \cite{ulyanov2016instance} and leaky ReLU activation. Both of the feature decoders are mirrored structures of the encoder excepting the final sigmoid activation functions. Our model configuration is consistent for all of the three datasets. The sizes of hidden content feature $h^{c}$ and motion feature $h^{m}$ are both 128. We employ the ADAM optimiser \cite{kingma2014adam} and set $\lambda_{1} = 1$, $\lambda_{3} = 1$ and $\alpha = 1$ for all models. Both the LSTM network for optical flow feature prediction and the Bi-direction LSTM for shuffle discriminator consist of two layers with 64 hidden units. Besides, all optical flow images are generated by pre-trained PWCNet \cite{sun2018pwcnet} model.

\subsection{Results on Synthetic dataset}
One of the popular video prediction benchmarks is the Moving MNIST dataset that contains 10000 sequences, each of which has 20 frames in length showing two digits moving in the fame size of 64 $\times$ 64. It has been widely used in recent video prediction works \cite{srivastava2015uselstm,denton2017drnet,oliu2018folded}. We follow the same experimental setting that use the first ten frames to predict the last ten in the video. For training, we set learning rate as $\num{1e-5}$, $\lambda_{2} = 0.01$, $\lambda_{4} = 0.01$ and $\beta = \num{1e-5}$.

\begin{figure*}
\centering
  \includegraphics[width=0.9\textwidth]{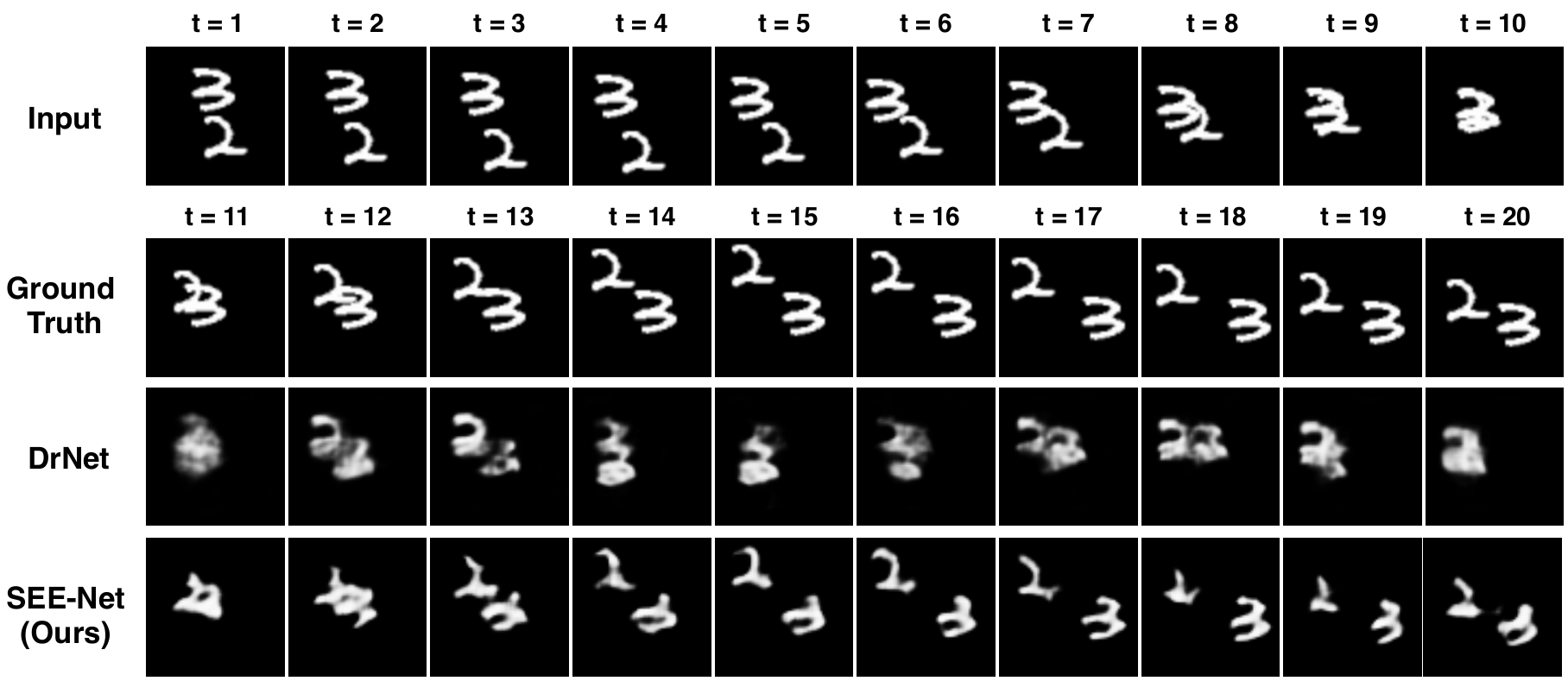}
  \caption{Qualitative Comparison to state-of-the-art methods on the Moving MNIST dataset. Using the first ten frames as input, the goal aims to predict the following 10 frames up to the end of the video. MCNet result is missing due to all the output images are black.}
  \label{fig:mnist pred}
\end{figure*}

Our major comparison is illustrated in Fig. \ref{fig:mnist pred}. It is noticeable that the results of MCNet are missing due to completely failed outputs with no contents generated. This result is consistent with existing reports that contain only quantitative MSE evaluation results. We attribute such failure to their temporal information is simply captured by frame difference. The disentangle representation in DrNet can force to preserve both the spatial and temporal information. In contrast, our explicit control over the generated sequence order can benefit the predicted digital numbers to be separated by the estimated movement trend from the sequential orders. It can be seen that the centre of each digital number aligns well with the ground truth.

\subsection{Results on Realistic Datasets}
KTH action dataset \cite{laptev2004KTH} and MSR action dataset \cite{li2010msr} are both used for evaluating video prediction methods. Actions in the traditional KTH dataset are simpler (walking, jogging, running, boxing, handwaving, handclapping) with more solid background therefore are suitable for predicting longer sequences, \textit{i.e.} using first 10 frames to predict the rest 20. Following \cite{villegas2017mcnet}, we apply person 1-16 for training and person 17-25 for test. The size of the input video is resized to 128 $\times$ 128. We set learning rate as $\num{1e-5}$, $\lambda_{2} = 1$, $\lambda_{4} = 1$ and $\beta = 0.001$ for training. MSR action dataset, in comparison, is closer to realistic scenarios with more cluttered background. We adopt a similar setting as that of KTH dataset and apply person 2 and person 6 for test. Following Mathieu \etal \cite{mathieu2015gdl}, the input frames is also 10 and the goal is to predict the rest 10 future frames.

\begin{figure*}
  \includegraphics[width=\textwidth]{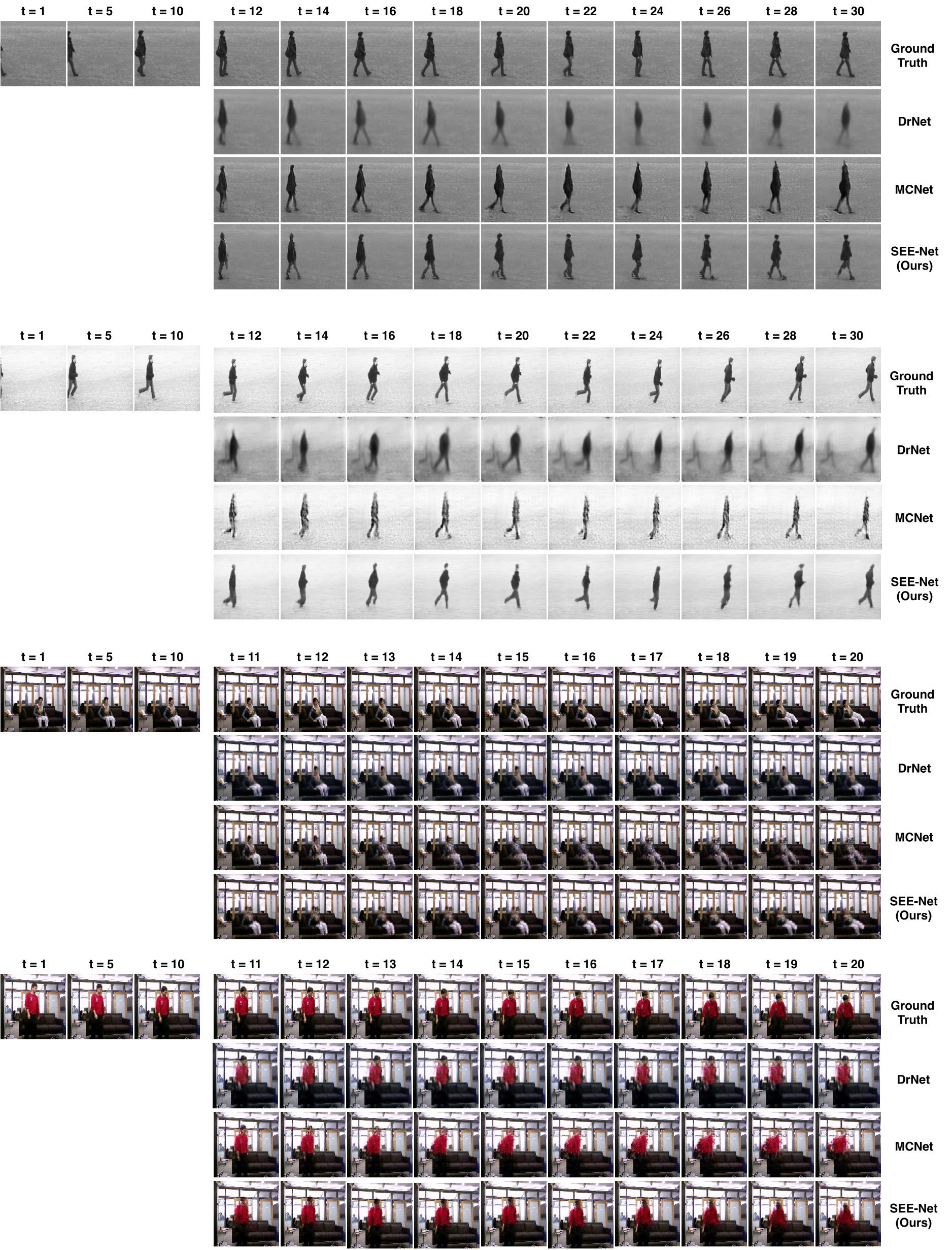}
  \caption{Qualitative comparison to state-of-the-art methods on KTH and MSR datasets. Due to the paper length, we only visualise input frame 1, 5 and 10 and predicted results on both datasets.}
  \label{fig:all pred}
\end{figure*}

In \figref{fig:all pred}, we can see that MCNet performs better than DrNet. On KTH dataset, DrNet often leads to loss of temporal information or distortion of the target. Although the content and motion information is forced preserved by the disentangled representation, the model fails to predict the correct movement trends between frames. MCNet can better match the trend of moving targets. However, it suffers from severe content information loss. The details of person and face are significantly distorted on both of the datasets. In contrast, our SEE-Net first demonstrates accurate trend perdition that can well synchronise with the ground truth movement. Meanwhile, our method can preserve more content details. We ascribe our success to the shuffle discriminator that can not only detect shuffled orders but also examine whether the generated features are realistic.

\begin{figure}
    \centering
    \begin{subfigure}[b]{\subfigurewidth}
        \includegraphics[scale=0.38]{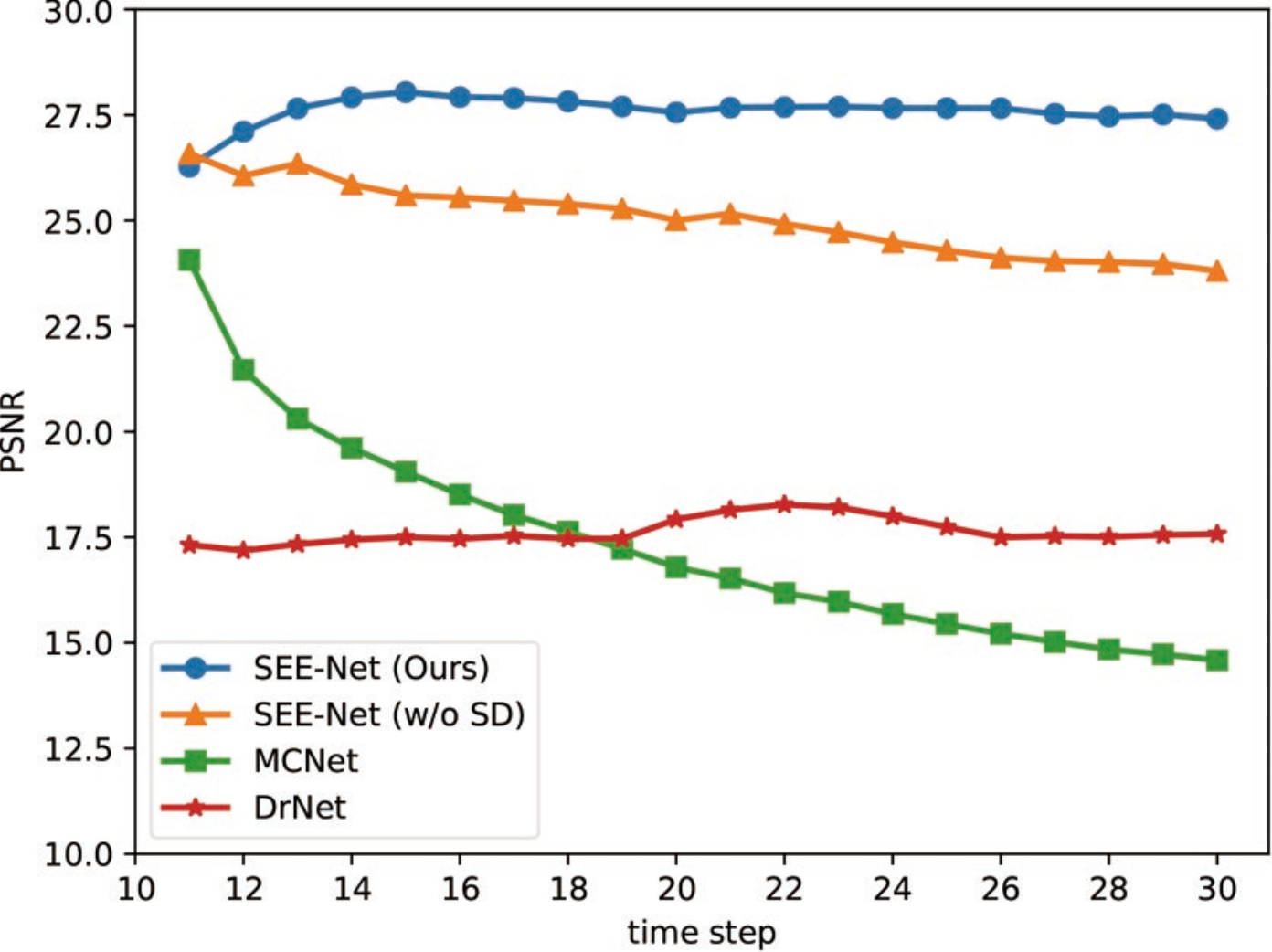}
        \caption{}
    \end{subfigure}%
    \begin{subfigure}[b]{\subfigurewidth}
        \includegraphics[scale=0.38]{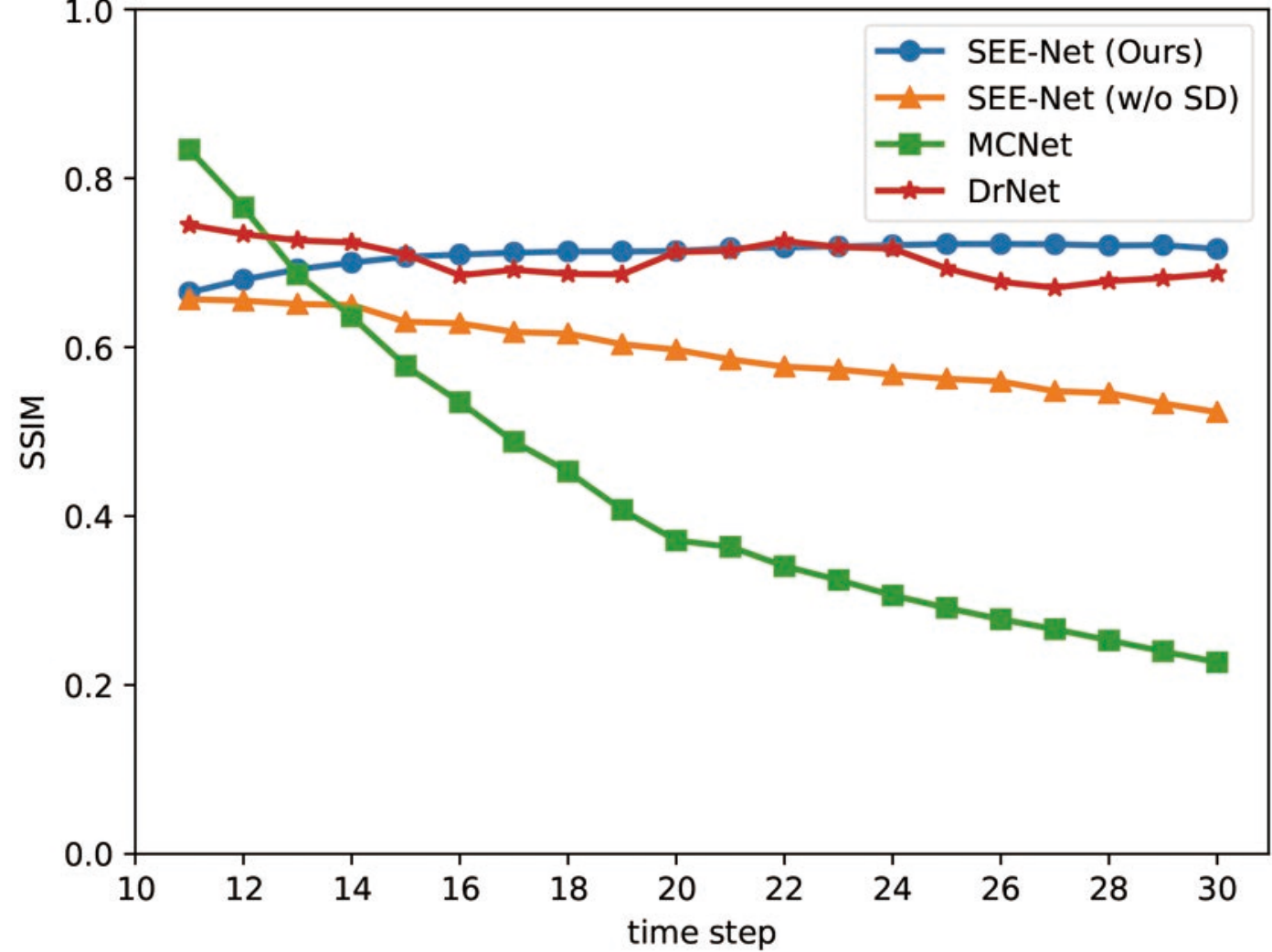}
        \caption{}
    \end{subfigure}%
    \newline
    \begin{subfigure}[b]{\subfigurewidth}
        \includegraphics[scale=0.38]{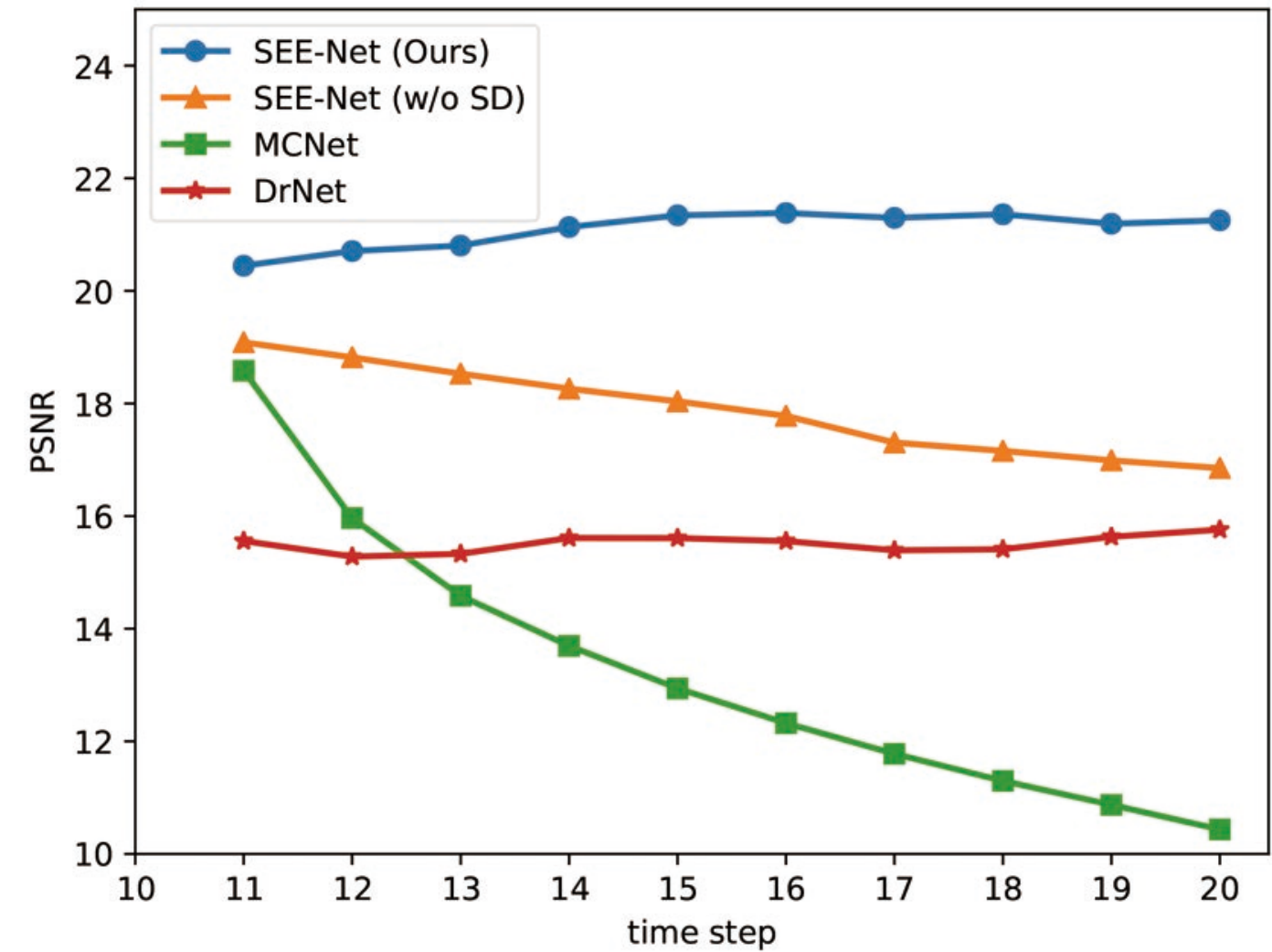}
        \caption{}
    \end{subfigure}%
    \begin{subfigure}[b]{\subfigurewidth}
        \includegraphics[scale=0.38]{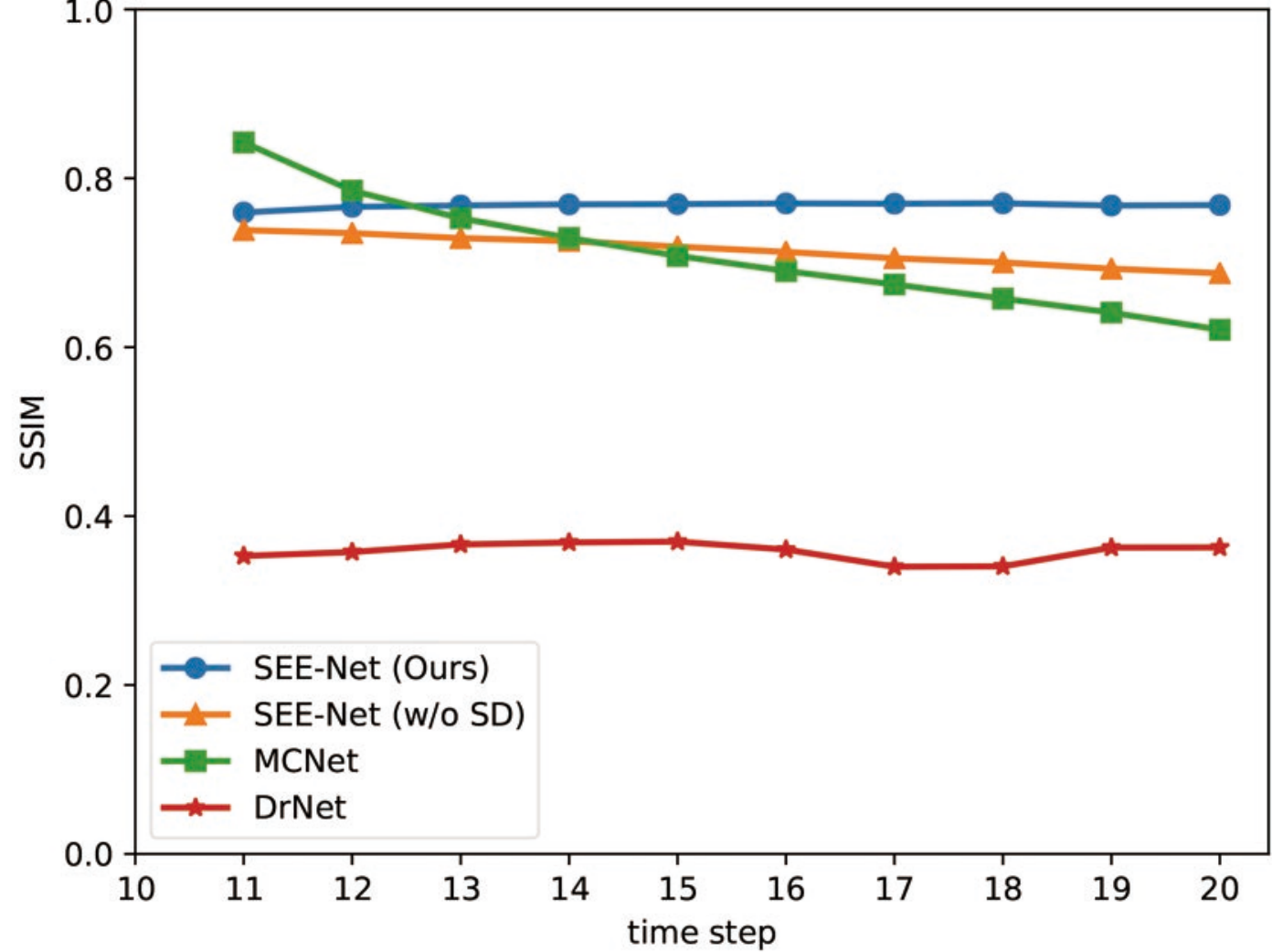}
        \caption{}
    \end{subfigure}%
    \caption{Quantitative results of PSNR and SSIM on KTH (a and b) and MSR (c and d) datasets. Compared with MCNet, DrNet, and our model without shuffling sequence.}\label{fig:animals}
    \vspace{-2ex}
\end{figure}

\begin{figure}
\centering
  \includegraphics[width=0.9\textwidth]{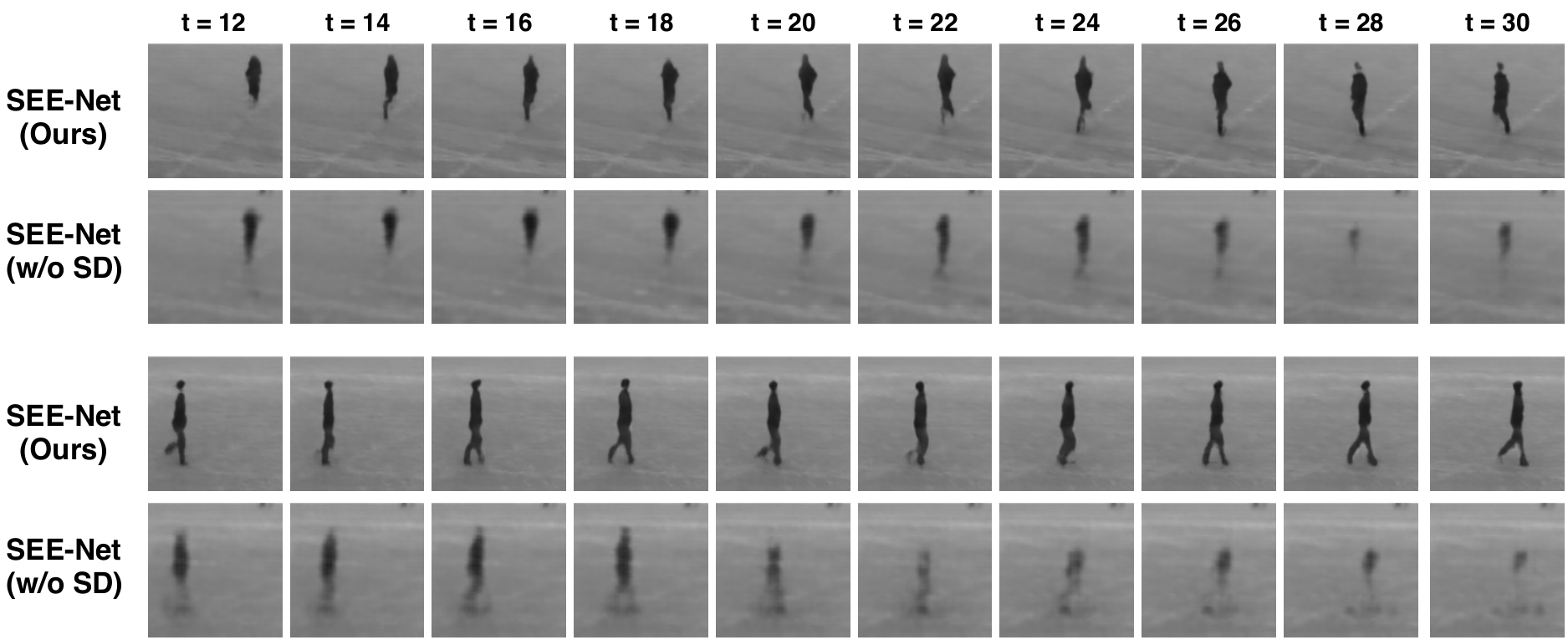}
  \caption{Comparing the predicted results on the KTH dataset based on the proposed SEE-Net and that without $SD$.}
  \label{fig:ab}
\end{figure}

In order to understand the overall performance on the whole dataset, we follow Mathieu \etal \cite{mathieu2015gdl} and employ Peak Signal to Noise Ratio (PSNR) and Structural Similarity (SSIM) as quantitative evaluation metrics in \figref{fig:animals}, from which we can draw similar conclusions as that from the qualitative comparison. Firstly, DrNet tends to preserve better content information and therefore achieves higher PSNR and SSIM scores compared to that of MCNet in KTH action dataset. In comparison, MCNet is more predictive so the scores are changing with the consecutive prediction. But its temporal information is gradually losing that results in severe detail loss of the generated frames. Our method consistently outperforms all of the compared approaches on both datasets and evaluation metrics.

It is also interesting to note that, in \figref{fig:animals}, the PSNRs and the SSIMs with respect to the SEENet (with or without $SD$) and the DrNet are more consistent than those of the MCNet. In our work, based on the assumption that a relatively stable background is provided, we fuse the content feature and the motion feature for future frame generation as described in \eqnref{eq:x^T+i}. Compared with MCNet, our approach can effectively avoid the accumulated error in the generation process.

\noindent\textbf{Effect of Shuffling Sequence Generation} To show the major contribution provided by our proposed shuffling method, we conduct an ablation study that removes the effect of shuffle discriminator by changing its hyper-weight to zero. Clear evidence can be found from the consecutive generated frames in \figref{fig:ab}, where the predicted target gradually resolves due to loss of temporal information. The powerful temporal and content information extracted by the consistency loss and optical flow network can retain the predictive power. However, the sequential information is not enhanced by discriminating the frame orders. Therefore, its overall performance is not as good as SEENet with shuffle discriminator.

\section{Conclusion}
This paper investigated shuffling sequence generation for video prediction using a proposed SEE-Net. In order to discriminate natural order from shuffled ones, sequential order information was forced to be extracted. The introduced consistency loss and optical flow network effectively disentangled content and motion information so as to better support the shuffle discriminator. On both synthesised dataset and realistic datasets, SEE-Net consistently achieved improved performance over state-of-the-art approaches. The in-depth analysis showed the effect of shuffling sequence generation in preserving long-term temporal information. In addition to the improved visual effects, SEE-Net demonstrated more accurate target prediction accuracy in both qualitative and quantitative evaluations. One of the important future work directions is to investigate new alternatives to reconstruction loss because it leads to averaged output in a long-term process. Another issue to be solved is the computational cost that limits the sequence prediction length at training stages.

\section{Acknowledgements}
Bingzhang Hu and Yu Guan are supported by Engineering and Physical Sciences Research Council (EPSRC) Project CRITiCaL: Combatting cRiminals In The CLoud (EP/M020576/1). Yang Long is supported by Medical Research Council (MRC) Fellowship (MR/S003916/1).

\bibliography{vpgan.bib}
\end{document}